\documentclass[10pt, conference, compsocconf]{IEEEtran}

\pdfoutput=1

\usepackage{cite}

\ifCLASSINFOpdf
  \usepackage[pdftex]{graphicx}
  \graphicspath{{content/figures}}
  \DeclareGraphicsExtensions{.pdf,.jpeg,.png}
\else
\fi

\usepackage[cmex10]{amsmath}
\usepackage{amsfonts}

\usepackage{array}
\usepackage{mdwmath}
\usepackage{mdwtab}
\usepackage{makecell}

\usepackage[caption=false,font=footnotesize]{subfig}

\usepackage{url}
\usepackage{hyperref}

\usepackage{placeins}

\begin{document}

\title{ProPanDL: A Modular Architecture for Uncertainty-Aware Panoptic Segmentation}

\author{\IEEEauthorblockN{Jacob Deery \ \ \ \ Chang Won Lee \ \ \ \ Steven L. Waslander}
\IEEEauthorblockA{Institute for Aerospace Studies \\
University of Toronto \\
Toronto, Canada \\
\{jacob.deery, changwon.lee\}@mail.utoronto.ca, steven.waslander@robotics.utias.utoronto.ca }
}

\maketitle

\begin{abstract}
We introduce ProPanDL, a family of networks capable of uncertainty-aware panoptic segmentation. Unlike existing segmentation methods, ProPanDL is capable of estimating full probability distributions for both the semantic and spatial aspects of panoptic segmentation. We implement and evaluate ProPanDL variants capable of estimating both parametric (Variance Network) and parameter-free (SampleNet) distributions quantifying pixel-wise spatial uncertainty. We couple these approaches with two methods (Temperature Scaling and Evidential Deep Learning) for semantic uncertainty estimation. To evaluate the uncertainty-aware panoptic segmentation task, we address limitations with existing approaches by proposing new metrics that enable separate evaluation of spatial and semantic uncertainty. We additionally propose the use of the energy score, a proper scoring rule, for more robust evaluation of spatial output distributions. Using these metrics, we conduct an extensive evaluation of ProPanDL variants. Our results demonstrate that ProPanDL is capable of estimating well-calibrated and meaningful output distributions while still retaining strong performance on the base panoptic segmentation task.
\end{abstract}


\section{Introduction}

The desire for full scene explainability in computer vision has spurred interest in \textit{panoptic segmentation} \cite{kirillov2019panoptic}. This task, requiring semantic segmentation of all pixels and instance discrimination for foreground pixels, unifies semantic and instance segmentation. Panoptic segmentation is appealing in that it provides theoretically full scene explainability: every pixel receives a predicted class, and every object is detected with the finest possible representation (pixel-level masks). However, at present, most panoptic and instance segmentation methods are capable of producing only point estimates: a single class and a single instance per pixel.

In parallel with research on panoptic segmentation, there has been a growing interest in \textit{uncertainty-aware deep learning} \cite{Gal2016Uncertainty}. An uncertainty-aware network outputs a full distribution over its predictions that implicitly captures their uncertainty, rather than a single point estimate. Applications for uncertainty estimation are numerous: well-calibrated uncertainty estimates can be used for anomaly detection \cite{Blum2021Fishyscapes}, offline for active learning \cite{Ebrahimi2020Uncertainty-guided}, or passed directly to consumers such as tracking or localization pipelines \cite{Akai2020Semantic}, which are often based on classical, probabilistic algorithms that can easily leverage predicted probability distributions.

\begin{figure}[!t]
\centering
\includegraphics[width=\linewidth]{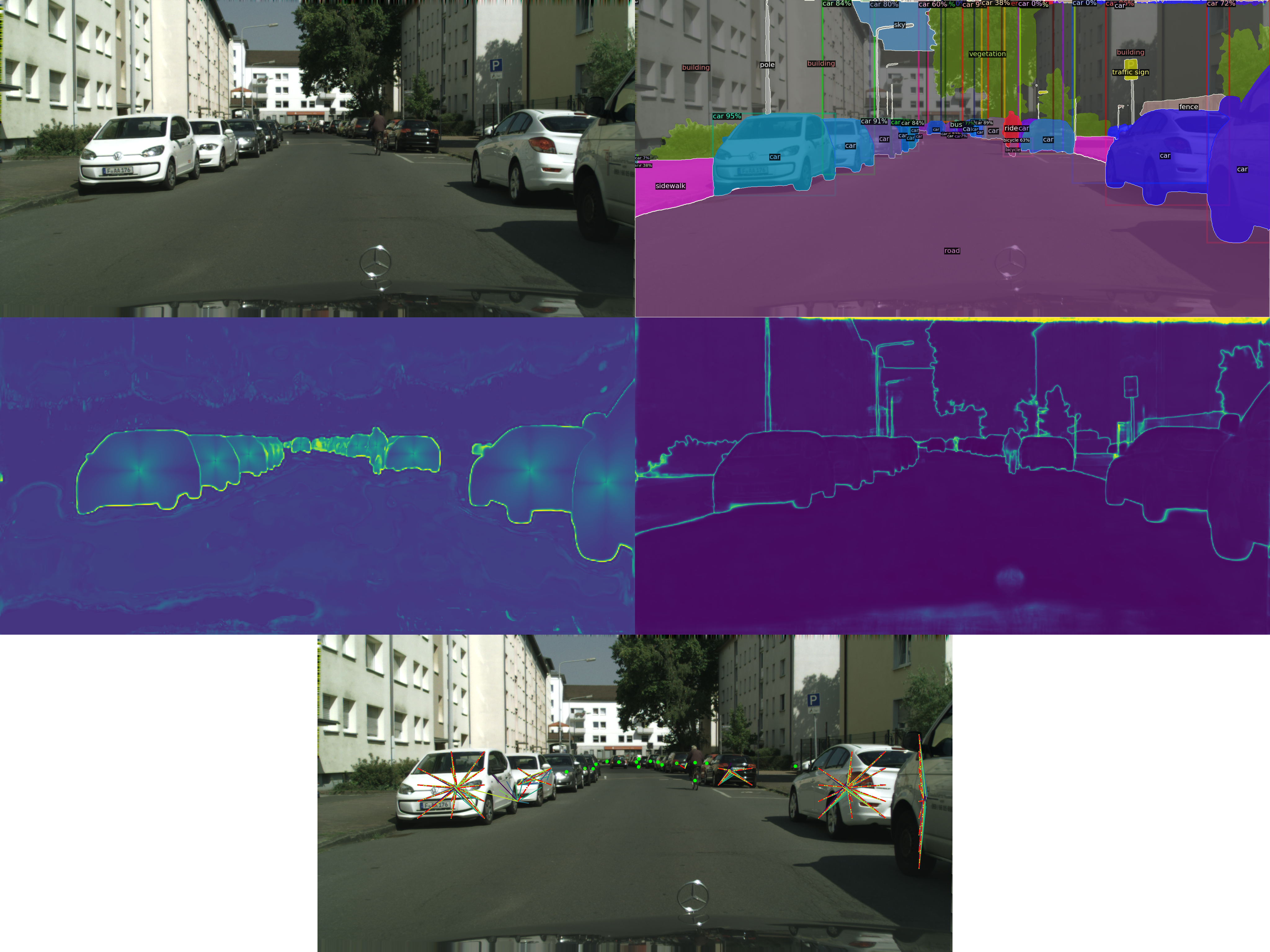}
\caption{ProPanDL outputs on a sample frame from the Cityscapes dataset. Clockwise from top left: input image, panoptic segmentation, semantic uncertainty, offset vector samples, spatial uncertainty.}
\label{fig:offset_example}
\end{figure}

To date, there has been little research exploring the application of uncertainty estimation to the panoptic segmentation task. Uncertainty-aware panoptic segmentation was introduced by Sirohi et al. \cite{sirohi2022uncertainty}, who introduce a network architecture capable of producing pixel-wise uncertainties as well two metrics, pECE and uECE, for evaluating the task. However, their approach is not without limitations. First, while panoptic segmentation has two aspects (semantic segmentation and instance discrimination), their proposed metrics evaluate only the total uncertainty rather than the two components independently. Second, their EvPSNet segmentation network is capable of only producing scalar uncertainty scores in the range $[0, 1]$, rather than full distributions. This limitation is reflected in the uECE and pECE metrics, which evaluate only real-valued uncertainty rather than full distributions.

To address these shortcomings, in this work, we introduce Probabilistic Panoptic-DeepLab (ProPanDL), a family of networks that extend the Panoptic-DeepLab architecture \cite{cheng2020panoptic} to produce pixel-wise predictive uncertainty in panoptic segmentation. To our knowledge, this work is the first to predict full output distributions for either instance or panoptic segmentation. ProPanDL uses separate semantic and spatial branches to jointly estimate semantic segmentation, instance discrimination, and their associated pixel-wise uncertainties. Semantic segmentation uncertainty is represented as a calibrated categorical distribution, produced via Temperature Scaling \cite{guo2017calibration} or Evidential Deep Learning \cite{sensoy2018evidential}. For instance discrimination, we apply Variance Networks \cite{nix1994estimating} and SampleNet \cite{harakeh2022estimating} to estimate the distribution of pixels relative to their corresponding instance center.

In addition, we propose to decouple the semantic and spatial uncertainty evaluation in pECE. Our proposed new metrics, $\text{pECE}_\text{spa}$ and $\text{pECE}_\text{sem}$, are straightforward modifications of the pECE metric adapted for separate spatial and semantic uncertainty evaluation. We additionally propose the use of the energy score \cite{gneiting2007strictly}, a proper scoring rule used for robust evaluation of regression predictive uncertainty \cite{harakeh2021estimating}, for evaluation of the distributions produced by the ProPanDL spatial branch.

We conduct an extensive evaluation of ProPanDL variants, evaluating on both the standard Panoptic Quality metric and the uncertainty-aware metrics. We find that SampleNet + Evidential Deep Learning is the strongest combination, capable of estimating reliable spatial and semantic distributions while still retaining strong segmentation perfomance.

In summary, our contributions are threefold:
\begin{enumerate}
    \item We introduce Probabilistic Panoptic-DeepLab (ProPanDL), a family of networks extending the Panoptic-DeepLab architecture \cite{cheng2020panoptic} for uncertainty-aware segmentation.
    \item We propose two metrics, $\text{pECE}_\text{spa}$ and $\text{pECE}_\text{sem}$, for separate evaluation of spatial and semantic uncertainty.
    \item We conduct a thorough evaluation of ProPanDL and analyze each proposed uncertainty estimation method, demonstrating that our approach is capable of robust panoptic uncertainty estimation.
\end{enumerate}

\section{Related Work}

\subsection{Panoptic Segmentation}

Panoptic segmentation methods can be categorized into one of two approaches. \textit{Top-down} methods merge the outputs of semantic segmentation and proposal-based instance segmentation branches. These methods require additional postprocessing to resolve overlaps between individual instances and semantic segmentation boundaries \cite{mohan2021efficientps} \cite{porzi2019seamless} \cite{xiong2019upsnet} \cite{kirillov2019fpn}. Conversely, \textit{bottom-up} methods predict non-overlapping segments directly.

At the time of writing, Panoptic-DeepLab (PDL) \cite{cheng2020panoptic} is the state of the art panoptic segmentation method, ranking at the top of the Cityscapes \cite{cordts2016cityscapes} leaderboard. PDL is a bottom-up method that operates by predicting the centers of instance masks, regressing every pixel to a corresponding center, and classifying the resulting segments using a parallel semantic segmentation branch.

\subsection{Uncertainty Estimation}

\textit{Uncertainty-aware deep learning} is generally used to refer to any methods which extend deep learning to estimation of full probability distributions rather than point estimates. Uncertainty is often classified into \textit{aleatoric} and \textit{epistemic} uncertainty. Epistemic uncertainty refers to the uncertainty present in the parameters of a neural network, and can be reduced arbitrarily by training the network with additional data. Conversely, aleatoric uncertainty is inherent in the data (e.g. sensor noise, quantization) and cannot be reduced via additional training.

\subsubsection{Uncertainty in Semantic Segmentation}

Due to the nature of the segmentation task, which is structured as a pixel-wise classification problem, many of the uncertainty estimation methods developed for image classification can be extended directly to segmentation. The first work to investigate this application was performed by Kendall and Gal \cite{kendall2017uncertainties}, who applied Monte Carlo Dropout (MCD) \cite{gal2016dropout} in conjunction with direct variance estimation \cite{Gal2016Uncertainty} to jointly estimate aleatoric and epistemic uncertainty. MCD approximates Bayesian inference over the network weights by sampling the same input through a fully-trained network with Dropout enabled at inference time.

Due to its simplicity and strong performance, MCD is widely used \cite{mukhoti2018evaluating} \cite{mukhoti2021deterministic} \cite{isobe2017deep} as a baseline uncertainty estimation method. However, MCD requires multiple forward passes and is therefore computationally intensive and generally cannot meet latency requirements to run online in an autonomous driving setting. For this reason, there has been increasing interest in "direct modelling" methods, which directly predict the parameters of a probability distribution in one forward pass of the network \cite{feng2021review}. As the target output in classification is a categorical distribution, the Maximum Softmax Probability has been proposed as a simple baseline for confidence estimation \cite{hendrycks17baseline}, with temperature scaling a common variant to improve network calibration \cite{guo2017calibration}. Other work investigates regressing a distribution-over-distributions to quantify the network's confidence in the predicted class probabilities. Both \cite{sensoy2018evidential} (Evidential Deep Learning) and \cite{malinin2018predictive} (Dirichlet Prior Networks) adopt this approach, drawing from Dempster-Shafer evidential reasoning with the Dirichlet distribution as the target. The major difference between the methods is the use of out-of-distribution data for regularization in \cite{malinin2018predictive}.

\subsubsection{Uncertainty in Instance Segmentation}

Unlike semantic segmentation, few published works explore uncertainty estimation for instance segmentation. Morrison et al. \cite{morrison2019uncertainty} obtain instance-level uncertainty estimates by applying Monte Carlo Dropout sampling to Mask R-CNN. Their method is instance-centric and is not capable of estimating pixel-wise uncertainty. Rumberger et al. \cite{rumberger2020probabilistic} use a similar dropout-based approach, but employ a bottom-up metric learning approach to produce overlap-free masks and pixel-wise uncertainties. Both of these approaches rely on dropout sampling and are therefore ill-suited to deployment in autonomous driving applications.

\subsubsection{Panoptic Segmentation}

Sirohi et al. introduce the uncertainty-aware panoptic segmentation task in \cite{sirohi2022uncertainty} which, to the best of our knowledge, remains the only work on this task. The authors introduce two new metrics: the uncertainty-aware calibration error uECE and the panoptic calibration error pECE. Using the EfficientPS network \cite{mohan2021efficientps} as a base network, the authors propose the new EvPSNet architecture, applying evidential deep learning separately to the semantic segmentation and instance prediction branches. The predicted semantic and instance segmentation map and masks and their associated uncertainties are fused by a post-processing panoptic fusion module to produce pixel-wise semantic class predictions, instance IDs, and uncertainties. While this method estimates semantic and instance uncertainty separately, it fuses them into a total panoptic uncertainty map. We argue that separately estimating and evaluating each component of uncertainty is valuable; downstream tasks, such as tracking or localization, may be able to benefit greatly from knowing that the location is certain but the class is invalid, or vice versa. Moreover, while EvPSNet estimates only a single real-valued uncertainty for every pixel, our method is capable of estimating full posterior distributions for both semantic and instance predictions, which provide more information to downstream consumers.

\subsubsection{Regression Tasks}

The most common approaches to uncertainty estimation in regression assume some parametric distribution for the output and modify the network to estimate the parameters of this distribution. Gaussian (Variance Networks \cite{nix1994estimating}) or Mixture of Gaussians (Mixture Density Networks \cite{bishop1994mixture}) are common assumptions and have been widely used for bounding box uncertainty in probabilistic object detection \cite{harakeh2020bayesod} \cite{he2020deep} \cite{feng2021review} and dense regression tasks such as depth prediction \cite{tosi2021smd}.

More recently, Harakeh et al. \cite{harakeh2022estimating} introduced SampleNet, a network architecture for estimating parameter-free distributions. By modifying the last layer of a network to estimate $M$ samples rather than a single point estimate, the network can accurately model distributions for which parametric assumptions, such as Gaussian, may not hold. The network requires only one forward pass during both training and inference to produce all $M$ samples, resulting in minimal additional overhead. The authors successfully applied SampleNet to monocular depth estimation; we adapt this approach for pixel offset regression in Panoptic-DeepLab, a similar dense regression task.

\subsubsection{Evaluation}

The widely accepted standard metrics for uncertainty evaluation are the Expected Calibration Error (ECE) and Maximum Calibration Error (MCE) \cite{guo2017calibration}. Sirohi et al. \cite{sirohi2022uncertainty} propose a modified form of the ECE which they call the uECE, as well as an additional pECE metric used specifically for panoptic segmentation. However, these metrics evaluate only the total uncertainty and do not separately evaluate semantic and spatial uncertainty. Moreover, they are incapable of evaluating the full probability distributions that many networks are capable of providing; calibration requires uncertainty to be represented as a single real number in the range $[0, 1]$. We propose a modified form of pECE that addresses the first limitation and employ the alternative energy score \cite{gneiting2007strictly} to address the second.

For evaluation of probability distributions, proper scoring rules are typically used. A scoring rule is \textit{proper} if it reaches its minimum when the evaluated distribution is identical to the target distribution \cite{harakeh2021estimating} and is called \textit{strictly proper} if it is greater for all other distributions. For classification, the Brier score \cite{brier1950verification} and standard cross-entropy loss are common proper scoring rules; for regression, the negative log-likelihood (NLL) and energy score (ES) \cite{gneiting2007strictly} are widely used \cite{harakeh2021estimating}. We use energy score both as a training loss for ProPanDL and as an evaluation metric of the pixel-wise probability distributions predicted by the offset regression branch.

\section{Methodology}

\begin{figure*}[!t]
\centering
\includegraphics[width=\linewidth]{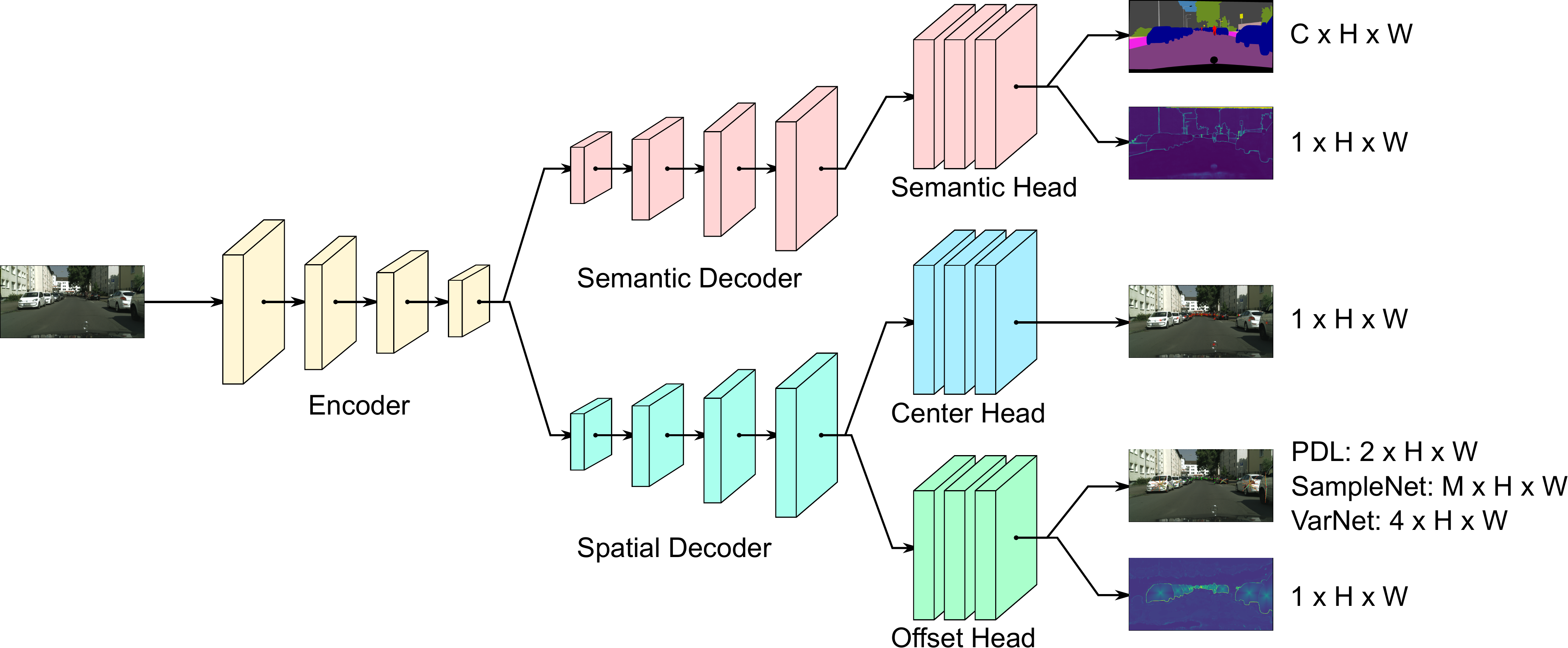}
\caption{An overview of the ProPanDL network architecture. The shared encoder backbone, semantic decoder, spatial decoder, and center head are unchanged from Panoptic-DeepLab \cite{cheng2020panoptic}. We modify the semantic head to produce calibrated pixel-wise categorical distributions and uncertainty estimates. Similarly, we modify the offset head to predict a distribution of pixel offset vectors rather than point estimates.}
\label{fig:propandl}
\end{figure*}

The task of panoptic segmentation involves two distinct subtasks: pixel classification and instance discrimination. It is logical and intuitive that any uncertainty method should be capable of quantifying uncertainty separately for each aspect. We call the uncertainty associated with semantic segmentation the \textit{semantic} uncertainty and the uncertainty associated with instance discrimination the \textit{spatial} uncertainty, mirroring the parallel terms used in probabilistic object detection \cite{hall2020probabilistic}.

We select Panoptic-DeepLab (PDL) \cite{cheng2020panoptic} as the meta-architecture for our family of uncertainty estimation networks, which we call Probabilistic Panoptic-DeepLab (ProPanDL). With separate branches for semantic segmentation and instance discrimination, PDL can be easily modified to produce separate semantic and spatial uncertainty maps.  The standard semantic segmentation branch follows the DeepLabV3+ decoder architecture \cite{chen2018encoder} and produces a $C$-dimensional vector for each pixel indicating the predicted class probabilities. The spatial branch of PDL produces two maps: the center prediction map and the center offset regression map. The center prediction map is a heatmap that predicts whether each pixel is located at the centroid of an instance, while the center offset regression map regresses the $x, y$ offset from each pixel to the center of the instance to which it belongs. 

In order to produce the final panoptic segmentation map, PDL employs a simple postprocessing stage. First, all predicted \textit{stuff} pixels are copied directly from the semantic segmentation map. Second, thresholding and non-max suppression are applied to the center prediction heatmap to produce individual predicted instance centers. Third, all \textit{thing} pixels are associated to the instance center closest to their predicted offset. All pixels associated to one center are considered to be part of the same segment. Fourth, each segment is given a semantic class based on simple majority vote of the pixels it contains. This results in a single class assigned to each pixel and a single instance ID to each \textit{thing} pixel.

In order to enable uncertainty-aware panoptic segmentation, our ProPanDL modifies the semantic head and pixel offset head of PDL. The modular nature of this approach enables us to test different spatial and semantic uncertainty methods in different combinations. When equipped for both spatial and semantic uncertainty, ProPanDL is capable of estimating pixel-wise calibrated categorical distributions in the semantic branch and pixel-wise multivariate distributions of offset vectors in the spatial branch. An overview of the ProPanDL architecture can be seen in Figure~\ref{fig:propandl}.

The following sections detail the uncertainty estimation methods we apply to each branch. In all equations, we use $i$ to denote an individual pixel, $N$ for the number of pixels in an image, and $C$ for the number of semantic classes.

\subsection{Semantic Uncertainty Estimation}

\subsubsection{Temperature Scaling}

While the maximum class probability (MCP) or maximum softmax probability (MSP) is a widely used heuristic for classification confidence, modern neural networks tend to be overconfident. Temperature Scaling is a simple method proposed by Guo et al. \cite{guo2017calibration} for improving the calibration of a classifier that uses the softmax function as its last operation. A single parameter $T$, known as the temperature, is used to adjust the logits $\mathbf{z}$ before they are passed through softmax:
\begin{equation}
    \text{softmax}(\mathbf{z}) = \frac{e^{\mathbf{z}/T}}{\sum_{c=1}^{C}e^{z_c/T}}
\end{equation}
Since the same temperature is applied to all of the class logits, the maximum element does not change; therefore, temperature scaling only affects calibration and does not change the predicted class. The temperature value is usually treated as a learned parameter and is tuned by minimizing the cross entropy loss on the validation set. During this procedure, the pretrained weights of the network remain unchanged. We perform temperature scaling by training over 25 epochs with a learning rate of 0.001.

\subsubsection{Evidential Deep Learning}

Evidential Deep Learning \cite{sensoy2018evidential} is a sampling-free direct modeling method that can be easily applied to any network architecture. This approach requires modifications only to the last layer of the network. Rather than estimating class probabilities directly, the network is modified to instead regress the parameters $\alpha_{1 \hdots C}$ of a $C$-dimensional Dirichlet distribution. As these parameters have a lower bound of 1, we append a final softplus nonlinearity to the classifier output and add 1 to the resulting values. While in the original image classification setting this estimation was performed once per image, we estimate one Dirichlet distribution per pixel $i$ in the image. Three auxiliary quantities can then be computed:
\begin{align}
S_i &= \sum_{c=1}^C \alpha_i^c \\
u_i &= C / S_i \\
\hat{p}_i^c &= \alpha_i^c / S_i
\end{align}
The first quantity is known as the strength of the distribution. A higher strength indicates higher confidence in the predicted per-class evidence. The corresponding uncertainty $u_i$ is inversely proportional to the strength and always falls in the range $(0, 1]$. The estimated class probabilities $\hat{p}_i^c$ can be calculated by dividing the corresponding Dirichlet parameter $\alpha_i^c$ by the distribution strength. Although the evidential formulation of this method provides a direct definition of uncertainty, previous work has observed that this quantity often provides worse estimates than using the maximum class probability or entropy of the resulting categorical distribution \cite{sensoy2018evidential} \cite{grannas2020real}. We evaluate and compare both approaches.

Rather than training with the standard classification cross-entropy loss, the authors of \cite{sensoy2018evidential} propose several different loss terms that can be used for EDL. In this work, we use the Type II Maximum Likelihood form of the loss, shown below:
\begin{equation}
\begin{aligned}
\mathcal{L}_{\text{EDL}} &= \sum_{i=1}^N \sum_{c=1}^C y_{i}^c \left(\log \left(S_i\right)-\log \left(\alpha_i^c\right)\right)
\end{aligned}
\end{equation}

We additionally apply the KL divergence regularization term, which penalizes the Kullback-Liebler divergence between two Dirichlet distributions $D$. We first define $\boldsymbol{\tilde{\boldsymbol{\alpha}}_i}$ as the vector of predicted Dirichlet parameters $\boldsymbol{\alpha}_i$ with the true class parameter $\alpha_{i}^c$ replaced with 1. $D_A = D(\mathbf{p}_{\mathbf{i}} \mid \tilde{\boldsymbol{\alpha}}_{\boldsymbol{i}})$ is then the Dirichlet distribution parameterized by $\boldsymbol{\tilde{\boldsymbol{\alpha}}_i}$. The loss term penalizes the KL divergence between $D_A$ and the Dirichlet distribution with uniform parameters, $D_1 = D\left(\mathbf{p}_i \mid\langle 1, \ldots, 1\rangle\right)$.
\begin{align}
\mathcal{L}_{\text{KL}} &= \lambda_t \sum_{i=1}^N K L\left[D\left(\mathbf{p}_{\mathbf{i}} \mid \tilde{\boldsymbol{\alpha}}_{\boldsymbol{i}}\right) \| D\left(\mathbf{p}_i \mid\langle 1, \ldots, 1\rangle\right)\right]
\end{align}
The Kullback-Liebler divergence $KL$ between the two Dirichlet distributions can be computed as follows, where $\Gamma: \mathbb{R} \rightarrow \mathbb{R}$ is the \textit{gamma} function and $\psi: \mathbb{R} \rightarrow \mathbb{R}$ is the \textit{digamma} function:
\begin{equation}
\begin{aligned}
K L[&D\left(\mathbf{p}_i \mid \tilde{\boldsymbol{\alpha}}_i\right) \| D\left(\mathbf{p}_i \mid \mathbf{1}\right)] \\[0.5em]
&=\log \left(\frac{\Gamma\left(\sum_{c=1}^C \tilde{\alpha}_i^c\right)}{\Gamma(C) \prod_{c=1}^C \Gamma\left(\tilde{\alpha}_{i k}\right)}\right) \\[0.5em]
& +\sum_{c=1}^C\left(\tilde{\alpha}_i^c-1\right)\left[\psi\left(\tilde{\alpha}_i^c\right)-\psi\left(\sum_{c=1}^C \tilde{\alpha}_{i c}\right)\right]
\end{aligned}
\end{equation}
Since a value of 1 indicates no evidence, this term incentivizes the network to remove all evidence for every class besides the true one. While the $\mathcal{L}_{\text{EDL}}$ loss previously described rewards this as well, the authors of \cite{sensoy2018evidential} found that the KL divergence term resulted in improved performance when the annealing coefficient $\lambda_t$ was scheduled correctly. For our experiments, we found $\lambda_t=\min (0.1, t / 60)$ to work effectively, similar to the values proposed by \cite{sirohi2022uncertainty}.

\subsection{Spatial Uncertainty Estimation}

Unlike the semantic segmentation subtask, which is pixel-wise classification, the offset estimation subtask is a pixel-wise regression problem. We seek to estimate, for every pixel, the distribution of offset vectors rather than a single point estimate. In this section, we use $v_i$ to denote the ground truth offset vector at pixel $i$ and use $\hat{v}_{i}$ to denote the predicted offset vector for this pixel.

\subsubsection{Variance Network}

Our first approach is to augment Panoptic-DeepLab with a Variance Network (VarNet) \cite{nix1994estimating}. VarNets assume that the output is drawn from a multivariate normal distribution and seek to estimate the mean and covariance of this distribution. In applying VarNet to ProPanDL, we assume a diagonal covariance matrix, a common assumption in similar settings \cite{harakeh2020bayesod} \cite{harakeh2022estimating}. We change the dimension of the last layer from $H \times W \times 2$ to $H \times W \times 4$, estimating both the mean offset vector and the diagonal elements of the covariance matrix. We train VarNet with the standard negative log-likelihood (NLL) loss \cite{nix1994estimating}:
\begin{equation}
\begin{aligned}
\mathcal{L}_{\text{NLL}} &= \sum_{i=1}^N \left( \mathcal{L}_{\text{NLL}, x}(i) + \mathcal{L}_{\text{NLL}, y}(i) \right) \\
\mathcal{L}_{\text{NLL}, x}(i) &= \frac{1}{2} \ln \left[\sigma^2\left(\hat{v}_{i,x}\right)\right] + \frac{\left[v_{i,x}-y\left(\hat{v}_{i,x}\right)\right]^2}{2 \hat{\sigma}_{i,x}^2\left(\hat{v}_{i,x}\right)} \\
\end{aligned}
\end{equation}
At inference time, we feed the estimated mean forward to the Panoptic-DeepLab postprocessing stage. In order to produce the real-valued uncertainty map for pECE evaluation, we calculate the total variance for each pixel (sum of the variance in $x$ and $y$) and divide by the maximum variance across the training set to scale the data between 0 and 1.

\subsubsection{SampleNet}

While normally distributed data is a common assumption in uncertainty-aware deep learning and is likely reasonable for the offset vectors of most pixels, we expect that it may not hold for certain difficult cases in offset regression. In particular, a pixel on the boundary of two instances will possess uncertainty not just about the precise location of an instance center but also about which instance it should be associated with. The true distribution in this case is likely to be multimodal. In order to estimate cases for which simple parametric distributions may be inadequate, we adapt SampleNet \cite{harakeh2022estimating}. Rather than predicting a single offset vector, we modify the last layer of the offset head to predict $M$ different offset vectors $\hat{v}_{i,1...M}$, which we treat as samples from the underlying distribution. We train SampleNet using the pixel-wise energy score loss, defined below for a single image with ground truth offset vectors $v_i$:
\begin{equation}
\begin{aligned}
\mathcal{L}_{\text{ES}} = \sum_{i=1}^N & \left(\frac{1}{M} \sum_{j=1}^M\left\|\hat{v}_{i, j}-{v}_i\right\|\right. \\
& \left.-\frac{1}{2 M^2} \sum_{j=1}^M \sum_{k=1}^M\left\|\hat{v}_{i, k}-\hat{v}_{i, j}\right\|\right)
\end{aligned}
\end{equation}

Due to memory constraints, we cap the number of samples at $M = 10$ for all experiments. In order to produce panoptic segmentation results at inference time, we calculate predicted centers for all sample vectors in the same way as Panoptic-DeepLab, and assign centers to pixels via majority among all samples. We reduce the set of samples to scalar uncertainties with a similar approach to VarNet: we calculate the sample variance for every pixel and divide by the maximum variance seen for every pixel in the training set.

\subsection{Evaluation Metrics}

Sirohi et al. \cite{sirohi2022uncertainty} introduce two metrics for evaluating uncertainty-aware panoptic segmentation, defined as follows:
\begin{align}
\text{uECE} &= \sum_{r=1}^R \frac{|B_r|}{N} |\text{acc}(B_r) - \text{conf}(B_r)|
\end{align}

\begin{align}
\text{pECE} &= \frac{1}{S} \left( \sum_{(f,g)} \text{uECE}(f, g) + \sum_{\tilde{f}} \text{uECE}(\tilde{f}) \right)
\end{align}

The uECE, a modification of the standard ECE, is calculated by partitioning pixels into $R$ equally-spaced bins based on confidence, where $\text{conf}(x) = 1 - \text{unc}(x)$. Within each bin $B_r$, $\text{acc}(B_r)$ is the average pixel accuracy and $\text{conf}(B_r)$ is the average pixel confidence.

While uECE is task-agnostic and can be applied to any uncertainty-aware problem, pECE is specific to the panoptic segmentation task and is calculated by averaging uECE over all $S$ predicted segments. For true positive segments $f$ associated with ground-truth segments $g$, $\text{acc}(i) = 1$ if the semantic class agrees with the ground truth and the pixel $i$ lies in both $f$ and $g$. For false positive segments $\tilde{x}$, $\text{acc}(i) = 0$ for all pixels.

By averaging uECE over all segments, pECE ensures that calibration of all pixels is evaluated. Moreover, averaging over segments implicitly places a higher weight on segments with fewer pixels. This emphasizes performance on small \textit{thing} instances such as pedestrians and distant vehicles, which networks often struggle to properly segment.

While these metrics are useful for evaluating the total uncertainty, they do not allow for independent evaluation of spatial and semantic uncertainties. We argue that evaluating each quantity separately provides much more insight into the performance of the network, especially for modular architectures like ProPanDL that support different uncertainty estimation methods in each branch. Therefore, we propose the following modified metrics:
\begin{align}
\text{uECE}_\text{spa} &= \sum_{r=1}^R \frac{|B_r|}{N} |\text{acc}_\text{spa}(B_r) - \text{conf}_\text{spa}(B_r)|
\end{align}
\begin{equation}
\begin{aligned}
\text{pECE}_\text{spa} = & \frac{1}{S} \Big( \sum_{(f,g)} \text{uECE}_\text{spa}(f, g) \\
& + \sum_{\tilde{f}} \text{uECE}_\text{spa}(\tilde{f}) \Big)
\end{aligned}
\end{equation}
\begin{align}
\text{uECE}_\text{sem} &= \sum_{r=1}^R \frac{|B_r|}{N} |\text{acc}_\text{sem}(B_r) - \text{conf}_\text{sem}(B_r)|
\end{align}
\begin{equation}
\begin{aligned}
\text{pECE}_\text{sem} = & \frac{1}{S} \Big( \sum_{(f,g)} \text{uECE}_\text{sem}(f, g) \\
& + \sum_{\tilde{f}} \text{uECE}_\text{sem}(\tilde{f}) \Big)
\end{aligned}
\end{equation}
As the equations suggest, we simply separate the uECE and pECE metrics into separate spatial and semantic metrics. Semantic accuracy $\text{acc}_\text{sem}(i) = 1$ if pixel $i$ is given the correct semantic label, regardless of its instance association. Likewise, spatial accuracy $\text{acc}_\text{spa}(i) = 1$ if pixel $i$ belongs to both a true positive segment $f$ and its associated ground truth segment $g$.

In order to evaluate against the pECE metric proposed by Sirohi et al. \cite{sirohi2022uncertainty}, we are required to compute a total uncertainty map for every pixel. We define the simple heuristic:
\begin{equation}
\text{unc} = \max(\text{unc}_\text{spa}, \text{unc}_\text{sem})
\end{equation}
for all methods. For methods which do not estimate spatial uncertainty, we set $\text{unc}_\text{spa} = 0$. We use the Maximum Class Probability \cite{hendrycks17baseline} as semantic uncertainty metric for all methods except EDL, where we use the evidential uncertainty defined in Eq.~(3).

A notable shortcoming of both the original and modified metrics is that they operate only on scalar-valued uncertainty maps where uncertainty is represented by a single real number. In contrast, ProPanDL seeks to model full distributions for both semantic class and pixel offset. While evaluating a categorical distribution via Shannon entropy or maximum softmax probability is a standard approach \cite{Gal2016Uncertainty}, reducing a sample-modelled distribution to a single real number discards much of the valuable information present in the samples. We therefore use the energy score as an additional metric to directly evaluate the quality of the predicted offset distributions.

\section{Experimental Results}

\begin{table*}[!t]
\renewcommand{\arraystretch}{1.3}
\caption{Panoptic segmentation performance on Cityscapes}
\label{tab:cityscapes}
\centering
\begin{tabular}{c|cc||cccc}
& \multicolumn{2}{c}{Uncertainty} & \multicolumn{4}{c}{Performance} \\
\hline
Method & Spatial & Semantic & PQ $\uparrow$ & pECE $\downarrow$ & $\text{pECE}_\text{spa}$ $\downarrow$ & $\text{pECE}_\text{sem}$ $\downarrow$ \\
\hline
Panoptic-DeepLab \cite{cheng2020panoptic} & $-$ & $-$ & 60.7 & 32.5 & 32.7 & 22.1 \\
EvPSNet \cite{sirohi2022uncertainty} & \checkmark & \checkmark & 63.7 & 19.3 & $-$ & $-$ \\
\hline
SampleNet Only & \checkmark & $-$ & 59.3 & 30.2 & 27.6 & 22.5 \\
VarNet Only & \checkmark & $-$ & 54.2 & 38.4 & 33.3 & 22.5 \\
EDL Only & $-$ & \checkmark & 56.1 & 25.2 & 33.0 & 18.0 \\
TS Only & $-$ & \checkmark & 60.7 & 29.3 & 32.7 & 20.1 \\
\hline
SampleNet + TS & \checkmark & \checkmark & 59.3 & 30.4 & 27.4 & 20.3 \\
VarNet + TS & \checkmark & \checkmark & 54.2 & 38.0 & 33.3 & 20.3 \\
SampleNet + EDL & \checkmark & \checkmark & 58.8 & 29.2 & 29.2 & 19.5 \\
VarNet + EDL & \checkmark & \checkmark & 48.0 & 35.1 & 32.8 & 17.3 \\
\hline
\end{tabular}
\end{table*}

\subsection{Training}

We use the Detectron2 \footnote{\url{https://github.com/facebookresearch/detectron2}} implementation of Panoptic-DeepLab with a ResNet-50 backbone as the starting point for our experiments. The energy score used to train SampleNet is implemented using the GeomLoss library \cite{feydy2019interpolating}. We initialize all experiments from the Detectron2 weights trained on Cityscapes with ImageNet pretraining on the backbone. We leave the backbone frozen during training for all experiments.

We use a batch size of 10 and crop of 512 $\times$ 1024 for all experiments due to memory constraints. We train using the Adam optimizer \cite{kingma2015adam} with a polynomial learning rate schedule and no weight decay.

We independently train and evaluate all combinations of \{SampleNet \cite{harakeh2022estimating}, VarNet \cite{nix1994estimating}\} and \{TS \cite{guo2017calibration}, EDL \cite{sensoy2018evidential}\}.

\subsection{Quantitative Results}

Table~\ref{tab:cityscapes} shows the results of all methods evaluated on the Cityscapes validation set.

In terms of panoptic segementation performance, we observe that all ProPanDL methods drop in PQ relative to the PDL baseline. We note that the original Panoptic-DeepLab paper finds a batch size of 32 and no crop to be optimal, which may explain some of the performance drops that we observe. While all methods fall short of the PQ achieved by EvPSNet in \cite{sirohi2022uncertainty}, we reiterate that our goal was not to develop a state-of-the-art panoptic segmentation network; rather, our objective was to demonstrate that it is possible to estimate output distributions while not overly compromising performance on the base task.

Performance on the uncertainty estimation task varies by method and by metric. In terms of semantic uncertainty, both TS and EDL outperform the PDL baseline, with EDL outperforming TS in all settings. In terms of spatial uncertainty, SampleNet consistently outperforms VarNet. We further evaluate and analyze the spatial uncertainty methods in the following section.

We note that none of our methods perform similarly to EvPSNet on the pECE metric. We conjecture that this is due to the simple maximum heuristic which we use to calculate the pixelwise total uncertainty, in contrast to the panoptic fusion postprocessing stage which they adopt. However, as described earlier, we believe that scalar-valued uncertainty maps are inadequate for capturing all of the relevant uncertainty information, especially when semantic and scalar uncertainty are conflated.

\subsubsection{Offset distribution evaluation}

\begin{table}[!t]
\renewcommand{\arraystretch}{1.3}
\caption{Offset head characteristics}
\label{tab:rmse}
\centering
\begin{tabular}{c||cc|c}
Method & Avg. length & RMSE $\downarrow$ & ES $\downarrow$ \\
\hline
Ground truth & 79.04 & $-$ & $-$ \\
Panoptic-DeepLab \cite{cheng2020panoptic} & 74.22 & 18.95 & 17.84 \\
SampleNet \cite{harakeh2022estimating} & 66.12 & 38.95 & 3.47 \\
VarNet \cite{nix1994estimating} & 32.49 & 65.51 & 46.84 \\
\hline
\end{tabular}
\end{table}

In Table~\ref{tab:cityscapes}, we can observe that while SampleNet variants perform well in terms of PQ, VarNet performance drops by over 10\% in the worst case. We believe that this is due to the tendency of variance networks to produce high variance when trained with NLL, as described in \cite{harakeh2021estimating} and observed when training variance networks for depth estimation \cite{harakeh2022estimating}. In order to evaluate the predicted offset probability distributions independently of the center and semantic predictions, we calculate the energy score across all pixels. We calculate ES for VarNet by sampling from each pixelwise multivariate Gaussian distribution, similarly to \cite{harakeh2022estimating}.

\begin{figure}[!t]
\centering
\includegraphics[width=\linewidth]{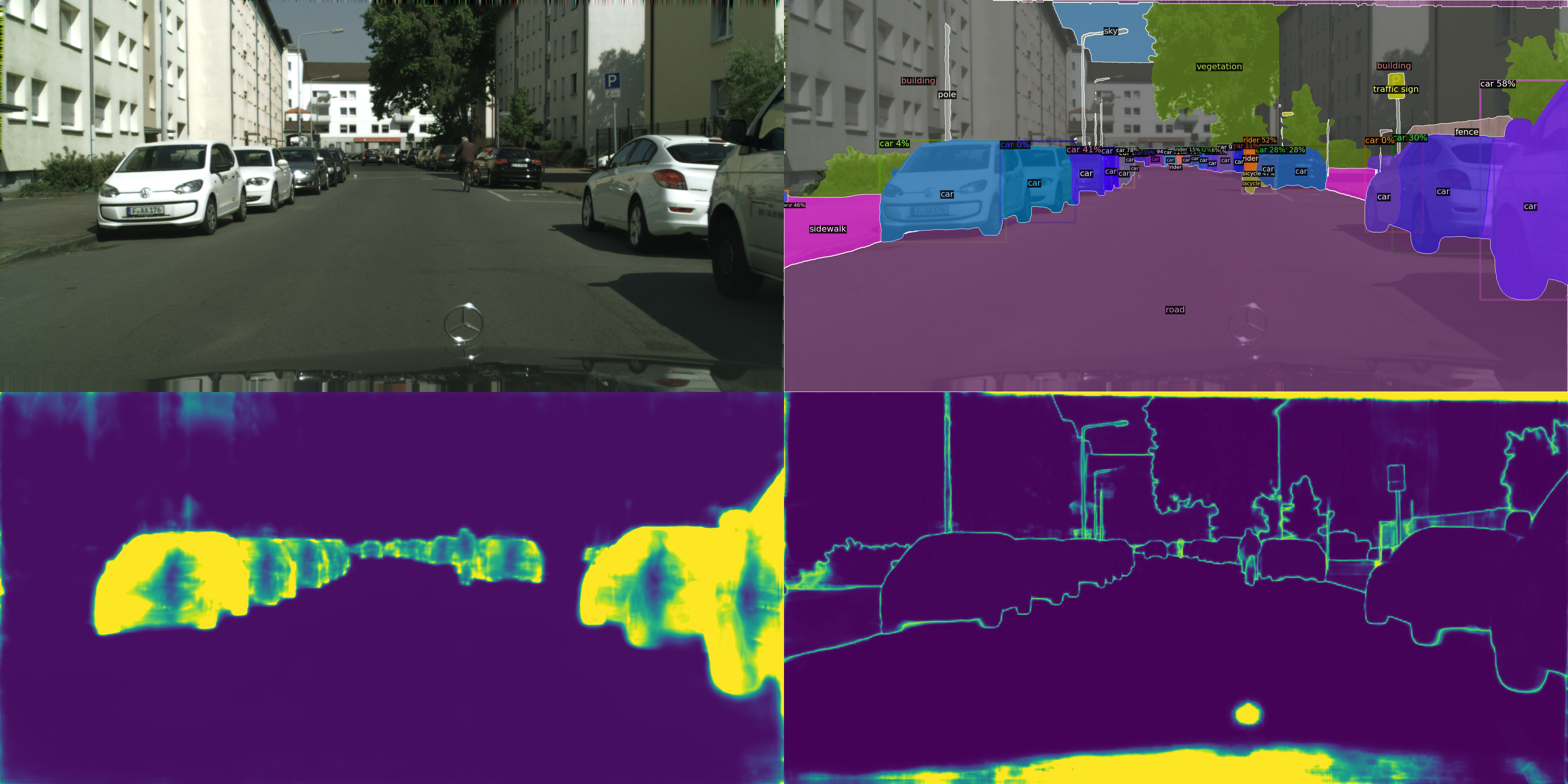}
\caption{An example figure produced by VarNet+EDL, demonstrating poor instance discrimination performance. Clockwise from top left: input image, panoptic segmentation, semantic uncertainty, predicted offset variance (spatial uncertainty). Brighter colours in uncertainty maps indicate higher uncertainty. While most "car" pixels are given the correct semantic class, we observe that the instance discrimination, as seen in the panoptic segmentation image, is poor.}
\label{fig:varnet_out}
\end{figure}

Table~\ref{tab:rmse} outlines the energy score for each model as well as the average length and RMS error of the predicted offset vectors. We observe that VarNet predicted sample vectors tend to be much shorter and have a significantly higher RMSE. Due to the postprocessing stage of Panoptic-DeepLab, pixels with shorter offset vectors are far more likely to end up clustered into an incorrect instance, which can result in significant PQ drops. Figure~\ref{fig:varnet_out} shows a representative example, where the predicted offset variance is highest inside vehicle instances, even in areas without overlap or ambiguity.

While SampleNet's RMSE is also notably worse than vanilla Panoptic-DeepLab, this is not unexpected; we calculate the RMSE over all samples, and the ES loss used by SampleNet gives it the freedom to predict samples that have high error in attempt to model the underlying distribution (i.e. multimodal predictions near instance boundaries).

In terms of energy score, we observe that SampleNet attains the lowest value, followed by PDL and then by VarNet. The high ES of VarNet is consistent with our observation that the predicted distributions appear to have very high variance. While NLL and ES are both proper scoring rules and thus attain their global minimum at the same value, their behaviour outside of the minimum is complementary; networks trained with NLL have a tendency to overestimate rather than underestimate the variance, which is harshly penalized by ES \cite{harakeh2021estimating}.

\subsubsection{Oracle test}

\begin{table}[!t]
\renewcommand{\arraystretch}{1.3}
\caption{Oracle study on ProPanDL performance}
\label{tab:oracle}
\centering
\begin{tabular}{ccc||ccc}
Offsets & Centers & Sem. seg & PQ & $\text{pECE}_\text{spa}$ & $\text{pECE}_\text{sem}$ \\
\hline
SampleNet \cite{harakeh2022estimating} & GT & EDL & 59.0 & 29.1 & 19.6 \\
SampleNet \cite{harakeh2022estimating} & PDL & GT & 79.4 & 20.3 & $-$ \\
SampleNet \cite{harakeh2022estimating} & GT & GT & 82.8 & 19.1 & $-$ \\
\hline
VarNet \cite{nix1994estimating} & GT & EDL & 48.6 & 33.2 & 17.5 \\
VarNet \cite{nix1994estimating} & PDL & GT & 74.0 & 33.4 & $-$ \\
VarNet \cite{nix1994estimating} & GT & GT & 78.2 & 36.4 & $-$ \\
\hline
PDL & PDL & PDL & 60.7 & 32.7 & 22.1 \\
PDL & GT & GT & 83.3 & $-$ & $-$ \\
\hline
\end{tabular}
\end{table}

We conduct an oracle study to investigate the impact of each panoptic segmentation aspect on the performance of ProPanDL. This is accomplished by providing ground truth instance centers and/or semantic segmentation during evaluation instead of the network predictions, and allows us to determine the performance upper bound when using different offset branches. The results can be seen in Table~\ref{tab:oracle}. We see that in all cases, using ground truth semantic segmentation provides the biggest increase in performance, a trend previously observed in \cite{cheng2020panoptic} and \cite{zhou2021panoptic}. We note that while SampleNet remains competitive with PDL, VarNet performance lags well behind, demonstrating the limitations of this method when applied to panoptic segmentation.

\section{Conclusion}

In this work, we introduced Probabilistic Panoptic-DeepLab, a family of networks capable of uncertainty-aware panoptic segmentation. ProPanDL variants model spatial and semantic uncertainty separately, producing calibrated distributions for every pixel in the image. Possible extensions to this work would leverage the rich information available in the predicted distributions and could include extension to panoptic tracking or uncertainty-guided domain adaptation. We hope this work motivates further research in the estimation and application of predictive uncertainty for panoptic segmentation.

\bibliographystyle{IEEEtran}
\bibliography{IEEEabrv, references}

\end{document}